\documentclass[conference]{IEEEtran}
\IEEEoverridecommandlockouts
\usepackage{cite}
\usepackage{amsmath,amssymb,amsfonts}
\usepackage{algorithmic}
\usepackage{graphicx}
\usepackage{textcomp}
\usepackage{xcolor}
\usepackage{fancyhdr}
\def\BibTeX{{\rm B\kern-.05em{\sc i\kern-.025em b}\kern-.08em
    T\kern-.1667em\lower.7ex\hbox{E}\kern-.125emX}}
\begin{document}

\title{Strategies to Counter Artificial Intelligence in Law Enforcement: Cross-Country Comparison of Citizens in Greece, Italy and Spain\\
\thanks{This project has received funding from the European Union’s Horizon 2020 research and innovation program under grant agreement No 883596 (AIDA - Artificial Intelligence and advanced Data Analytics for Law Enforcement Agencies). The information in this paper reflects only the authors' view and the Agency is not responsible for any use that may be made of the information it contains.}
}

\author{\IEEEauthorblockN{Petra Saskia Bayerl}
\IEEEauthorblockA{\textit{CENTRIC} \\\textit{Sheffield Hallam University}\\Sheffield, UK\\p.s.bayerl@shu.ac.uk}
\and
\IEEEauthorblockN{Babak Akhgar}
\IEEEauthorblockA{\textit{CENTRIC} \\\textit{Sheffield Hallam University}\\Sheffield, UK\\b.akhgar@shu.ac.uk}
\and
\IEEEauthorblockN{Ernesto La Mattina}
\IEEEauthorblockA{\textit{Engineering Ingegneria}\\\textit{Informatica}, Palermo, Italy\\ernesto.lamattina@eng.it}
\and
\IEEEauthorblockN{Barbara Pirillo}
\IEEEauthorblockA{\textit{Engineering Ingegneria}\\\textit{Informatica}, Palermo, Italy\\barbara.pirillo@eng.it}
\and
\IEEEauthorblockN{Ioana Cotoi}
\IEEEauthorblockA{\textit{Engineering Ingegneria}\\\textit{Informatica}, Palermo, Italy\\ioanacristina.cotoi@eng.it}
\and
\IEEEauthorblockN{Davide Ariu}
\IEEEauthorblockA{\textit{PLURIBUS ONE S.r.l.}\\Cagliari, Italy\\davide.ariu@pluribus-one.it}
\and
\IEEEauthorblockN{Matteo Mauri}
\IEEEauthorblockA{\textit{PLURIBUS ONE S.r.l.}\\Cagliari, Italy\\matteo.mauri@pluribus-one.it}
\and
\IEEEauthorblockN{Jorge García}
\IEEEauthorblockA{\textit{Vicomtech} \\Donostia-San Sebastián, Spain\\jgarciac@vicomtech.org}
\and
\IEEEauthorblockN{Dimitris Kavallieros}
\IEEEauthorblockA{\textit{Center for Security Studies (KEMEA)} and\\\textit{University of the Peloponnese}, Athens, Greece\\d.kavallieros@kemea-research.gr}
\and
\IEEEauthorblockN{Antonia Kardara}
\IEEEauthorblockA{\textit{Center for Security Studies (KEMEA)}\\Athens, Greece\\a.kardara@kemea-research.gr}
\and
\IEEEauthorblockN{Konstantina Karagiorgou}
\IEEEauthorblockA{\textit{Center for Security Studies (KEMEA)}\\Athens, Greece\\k.karagiorgou@kemea-research.gr}}

\maketitle

\begin{abstract}
This paper investigates citizens' counter-strategies to the use of Artificial Intelligence (AI) by law enforcement agencies (LEAs). Based on information from three countries (Greece, Italy and Spain) we demonstrate disparities in the likelihood of ten specific counter-strategies. We further identified factors that increase the propensity for counter-strategies. Our study provides an important new perspective to societal impacts of security-focused AI applications by illustrating the conscious, strategic choices by citizens when confronted with AI capabilities for LEAs. 
\end{abstract}


%==============================
\section{Introduction}
Artificial Intelligence (AI) is a critical asset for law enforcement agencies' (LEAs) efficiency and effectiveness, e.g., by optimizing the evidence gathering and analysis process in serious and organized crime cases or by aiding the discovery of new adversarial trends and malicious patterns. Simultaneously, there are legitimate concerns about their usage, chief amongst them that algorithms can reinforce social inequalities (e.g., with respect to minority groups or genders), lead to faulty decisions with dramatic real-life consequences and create inflexible, insensitive procedures that fail to take into account individuals’ unique circumstances yet cannot be challenged because the underlying rules are too complex or opaque \cite{leprietal18}.  

The knowledge of emerging LEA capabilities impact perceptions of security within societies, positively and negatively. These perceptions are linked to citizens’ decisions about online and offline behaviors such as risk-taking, sharing of personal information or use of privacy protection tools \cite{bayerlakhgar15,mit20}. Presently, discussions about side-effects of LEAs' capabilities emphasize potential chilling effects \cite{stoyetal20}. These discussions tend to ignore, however, that citizens are often very conscious and strategic in reacting to perceived privacy issues. For instance, protesters may don uniform clothing, goggles and face masks to try to avoid facial recognition software. Also, an increasing number of recommendations and tools emerge to obfuscate, `confuse' or even `weaponize own data' against data collection efforts \cite{bruntonnissenbaum16,mcguigan21,mit20}. 

This issue is urgent given the high degree of public concerns about privacy: e.g., 63\% of Americans think it is impossible to avoid government entities collecting data about them \cite{pew19} and over half decided against products or services because they worried about collection of personal information \cite{pew20}. Such changes in mass-behaviors have operational consequences for LEAs \cite{bayerlakhgar15}, including training and long-term viability of AI applications.

In this paper, we investigate citizens' counter-strategies, as part of a larger research project on AI capabilities for LEAs (AIDA, https://www.project-aida.eu/). The objective is to understand the extent and nature of reactions as strategic choices to manage privacy boundaries as one element of societal side-effects of AI usage in the security domain. This knowledge informs the design and deployment of AI capabilities, provides insights into their long-term viability and can support LEAs in their communication with the public about current and future AI applications. Country differences in public attitudes towards AI are well documented \cite{neudertetal20}. To ensure a comprehensive picture, we are thus collecting data across multiple countries. This paper describes first results from the three countries Greece, Italy and Spain. 

%==============================
\section{Methods}

%...........................................................
\subsection{Data collection and sample}
To collect attitudes towards LEA-based AI applications, we used an online survey in the three respective country languages. Invitations were distributed through websites and social media channels of authors' organizations. Across all three countries, 338 people reacted to the survey invitation. 81 reactions were excluded, either because participants did not provide any information (n=79) or failed to give informed consent (n=3). The remaining sample consisted of 256 participants, of which 97 stemmed from Italy, 81 from Spain and 78 from Greece.  62.5\% of participants were female. The majority of participants (61.3\%) held at least a bachelor or master degree, followed by PhD (12.9\%), secondary school (10.9\%)  and professional degree (3.5\%; rest: primary school, other or prefer not to say). The age distribution ranged from 10-34 years (41.0\%)  to 35-44 years (35.9\%) and 45-54 years (9.4\%) to 55 years or older (5.9\%) (rest: prefer not to say). Only a small percentage self-identified as member of an ethnic minority (3.1\%), while a larger group stated to have been victim of a crime in the past (37.9\%). 
Participants did not receive any renumeration or reward. The study was approved by the ethics committee of the first author's university.

%...........................................................
\subsection{Variables}
The survey was split into four parts capturing (1) attitudes towards AI usage by law enforcement, (2) strategies used to counter-act such AI applications, (3) attitudes towards police generally and (4) demographics and personal characteristics. For this paper, we focus on the two AI-specific parts. \emph{Attitudes towards AI usage by law enforcement} were measured using five items assessing benefits of AI technologies (e.g., ``The security of society depends on the use and development of such technologies as AI''; measured from 1:strongly disagree to 5:strongly agree). Exploratory factor analysis (EFA) validated that the five items belonged to the same factor both for the full sample as well as for each of the countries individually. They were thus combined into a single value for attitudes towards AI ($\alpha_{all}$ = .83; $\alpha_{per~country}$: .82-.85). \emph{Strategies used to counter-act AI applications} were measured by asking about the likelihood of ten specific behaviors (e.g., ``encrypt communication/emails'', ``stop visiting particular websites'', ``ask others not to post personal information about me''; measured from 1:extremely unlikely to 5:extremely likely). EFA for the whole sample resulted in one factor, while for individual countries two factors emerged. However, as in all cases the second factor only barely reached the threshold of 1.0 for squared loadings, we decided on theoretical grounds to combine the items into a single scale. Internal reliability of the scale was high with $\alpha_{all}$ = .90 (range: .86-.91). We further captured participants' knowledge about AI (1 item; 1:no knowledge at all ... 5:expert), their felt vulnerability to biased police decisions by AI (1 item; 1:not at all ... 5:extremely) and fear of crime (3 items, e.g., ``I am afraid to become a victim of a cybercrime'', 1:not at all ... 5:extremely, $\alpha_{all}$ = .78; $\alpha_{per~country}$: .76-.84) as potential influencing factors.

%==============================
\section{Results}

\subsection{Country comparison of counter-strategies}
The overall likelihood of counter-strategies in the whole sample was moderate (m = 2.72, sd = .94). Neither gender, age nor education level affected counter-strategies (ANOVAs), F(197,26) = 1.55, ns. However, as expected, likelihood differed significantly across countries, F(2,239) = 6.38, p$<$.01. While overall reactions in Italy and Spain were similar, Greek participants were significantly more likely to choose counter-strategies than participants from Spain (pairwise comparison, p = .002) and marginally more likely than Italian participants (p = .053). In contrast, countries did not differ with respect to general attitudes towards AI use by LEAs (m = 3.48, sd = .87), F(243,2) = 1.25, ns.

\begin{figure*}[t]
\centerline{\includegraphics[width=\textwidth]{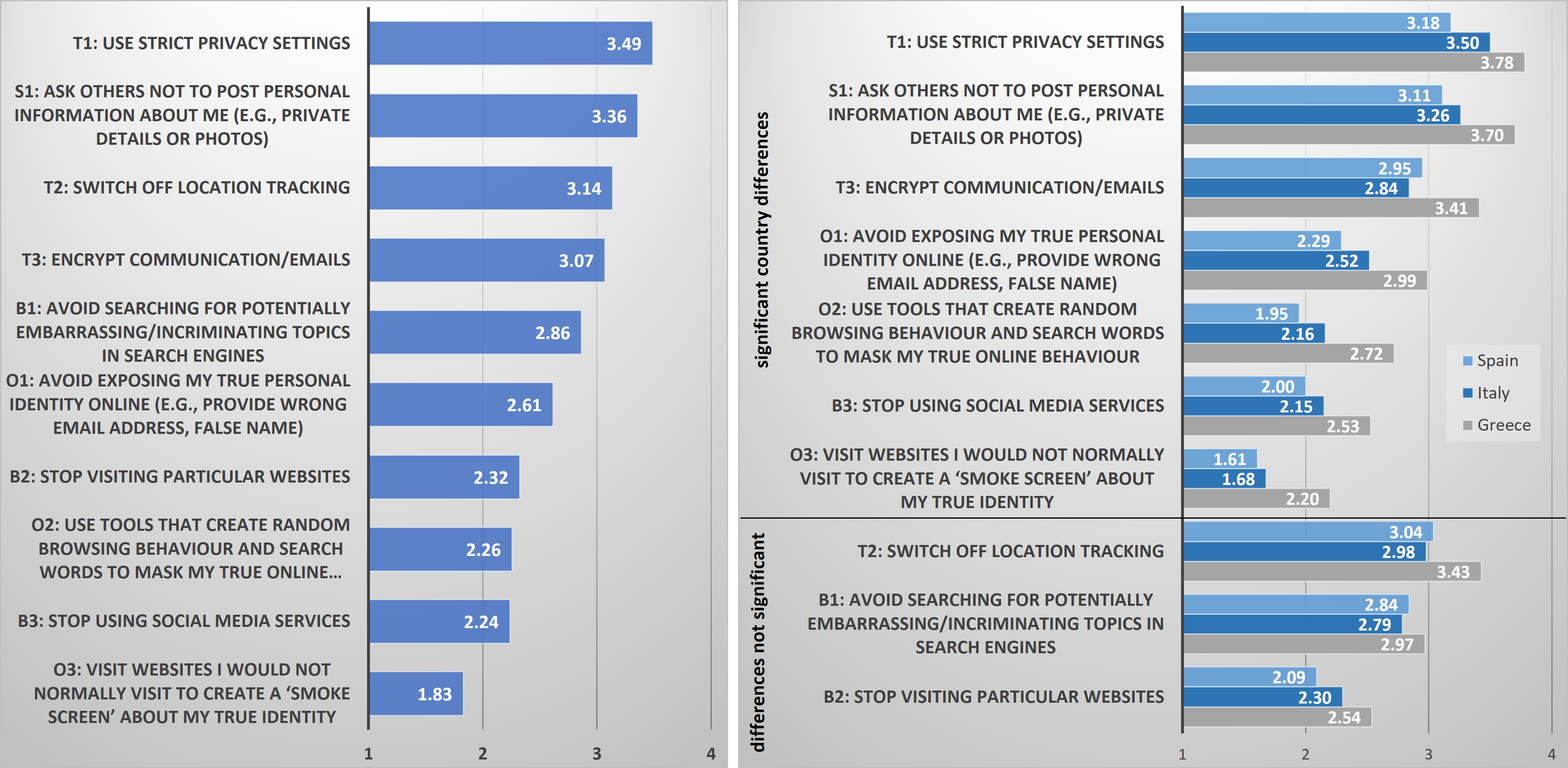}}
\caption{Individual counter-strategies: overall sample (left), country comparison (right)}
\label{fig1}
\end{figure*}

A closer look at individual counter-strategies illustrates important differences (cp. Fig. \ref{fig1}, left side). The primary strategies were technical: strict privacy settings, encryption and switching off location tracking (T1-T3). Next to this, participants favored asking others not to post (social strategy/S1) and to a lesser extent behavioral and obfuscation strategies of consciously not providing or giving false information (B1, O1). Other obfuscation strategies (O2: tools to create random browsing patterns or O3: creating a `smoke screen') were considered least likely. Participants were also very reluctant to stop using social media, despite the fact that social media often engender high levels of privacy concerns \cite{pew19}. Comparing countries shows that the four most prominent strategies (T1-T3 and S1) and the least favorite strategy (O3) were the same (cp. Fig. \ref{fig1}, right side). The remaining strategies showed some disparities (e.g., likelihood to stop using social media or use of obfuscation tools), however, only to a limited extent.

%...........................................................
\subsection{Influencing factors}
To investigate factors influencing counter-strategies, we ran a regression analysis with counter-strategies as dependent variable and countries (dummy coded), attitudes towards AI, perceived vulnerability to AI biases, fear of crime and AI knowledge as factors, using the enter method. The overall model was highly significant, F(229, 6) = 13.97, p$<$.001, $R^2$ = 0.27, adjusted $R^2$ = .25. It confirmed a higher level of counter-strategies for Greece. In addition, counter-strategies were significantly predicted by more critical attitudes towards AI use by LEAs, perceived vulnerability to AI biases and fear of crime (see table \ref{tab1}). In contrast, AI knowledge did not significantly affect counter-strategies.

\begin{table}[htbp]
\caption{Factors that affect counter-strategies}
\begin{center}
\begin{tabular}{|l|c|c|c|c|c|}
\hline
\textbf{Variable} & \textbf{B} & \textbf{SE B} & \textbf{$\beta$} & \textbf{t} & \textbf{p}  \\
\hline
(Constant) 						& 2.60  & 0.37 &          & 7.11  & .000\\
Country: Greece vs Italy   			& -0.13 & 0.13 & -0.07 & -0.96 & .336 \\ 
Country: Greece vs Spain 			& -0.43 & 0.14 & -0.21 & -3.19 & .002 \\ 
Attitudes AI use by LEAs			& -0.27 & 0.06 & -0.25 & -4.16 & .000 \\
Vulnerability to AI biases			& 0.25  & 0.05 & 0.30  & 5.05  & .000\\
Fear of crime 					& 0.17  & 0.06 & 0.17  & 2.96  & .003\\
AI knowledge 					& 0.08  & 0.06 & 0.09  & 1.49  & .137\\\hline
\end{tabular}
\label{tab1}
\end{center}
\end{table}

%==============================
\section{Discussion and next steps}
Our results illustrate a differentiated approach of citizens in their reaction to AI use by LEAs. Technical and social counter-strategies emerged as the most likely. In contrast, more radical obfuscation strategies and stopping the use of social media found little favor. Overall propensity for counter-strategies varied across countries. At the same time, there were considerable overlaps for the specific strategies, suggesting that the pattern of primary choices may be largely comparable. We further found that propensity for counter-strategies were grounded in personal attitudes towards LEAs' use of AI and perceived vulnerabilities to AI biases and crime. The findings support our assumption of conscious and strategic choices by citizens when confronted with AI use by LEAs. 

Our study expands the current discussion from primarily ethical and legal considerations towards concrete societal side-effects and their potential practical design and usage implications. With this it provides an important new perspective on societal impacts by illustrating the conscious, strategic citizen choices when confronted with LEAs' AI use. 

In the same regard, this study constitutes a preliminary exploration which clearly requires extension. The next steps will be a broader investigation of counter-strategies across the full set of eleven partner countries in AIDA as well as closer investigations about a broader set of counter-strategies within the four categories (technical, behavioral, social, obfuscation) and individuals' reasons for their choice.

\vspace{12pt}
\end{document}